\pdfoutput=1

\documentclass[11pt]{article}

\usepackage{acl}

\usepackage{times}
\usepackage{latexsym}

\usepackage[T1]{fontenc}

\usepackage[utf8]{inputenc}

\usepackage{microtype}
\usepackage{tabularx}
\usepackage{booktabs}       
\usepackage{amsfonts}       
\usepackage{amssymb}
\usepackage{nicefrac}       
\usepackage{xcolor}         
\usepackage{natbib}
\usepackage{amsmath}

\usepackage{graphicx}
\usepackage{subcaption}
\usepackage{enumitem}
\usepackage{multirow}
\usepackage{makecell}

\title{Data Augmentation for Intent Classification with Off-the-shelf Large Language Models}

\author{%
  Gaurav Sahu\thanks{Work done during an internship at ServiceNow Research} \\
  University of Waterloo\\
  \texttt{gsahu@uwaterloo.ca} \\
   \And
   Pau Rodriguez \\
   ServiceNow Research \\
   \And
   Issam H. Laradji \\
   ServiceNow Research \\
   \AND
   Parmida Atighehchian \\
   ServiceNow Research \\
   \And
   David Vazquez \\
   ServiceNow Research \\
   \And
   Dzmitry Bahdanau \\
   ServiceNow Research \\
}

\begin{document}

\maketitle

\begin{abstract}
    Data augmentation is a widely employed technique to alleviate the problem of data scarcity. In this work, we propose a prompting-based approach to generate labelled training data for intent classification with off-the-shelf language models (LMs) such as GPT-3.
    An advantage of this method is that no task-specific LM-fine-tuning for data generation is required; hence the method requires no hyper-parameter tuning and is applicable even when the available training data is very scarce.
    We evaluate the proposed method in a few-shot setting on four diverse intent classification tasks.
    We find that GPT-generated data significantly boosts the performance of intent classifiers when intents in consideration are sufficiently distinct from each other. In tasks with semantically close intents, we observe that the generated data is less helpful. Our analysis shows that this is because GPT often generates utterances that belong to a closely-related intent instead of the desired one. We present preliminary evidence that a prompting-based GPT classifier could be helpful in filtering the generated data to enhance its quality.\footnote{Our code is available at: \url{https://github.com/ElementAI/data-augmentation-with-llms}}
\end{abstract}

\section{Introduction}
A key challenge in creating task-oriented conversational agents is gathering and labelling training data.
Standard data gathering options include manual authoring and crowd-sourcing. Unfortunately, both of these options are tedious and expensive.
\textit{Data augmentation} is a widely used strategy to alleviate this problem of data acquisition.

\begin{figure}[t!]
    \centering
    \includegraphics[width=\linewidth]{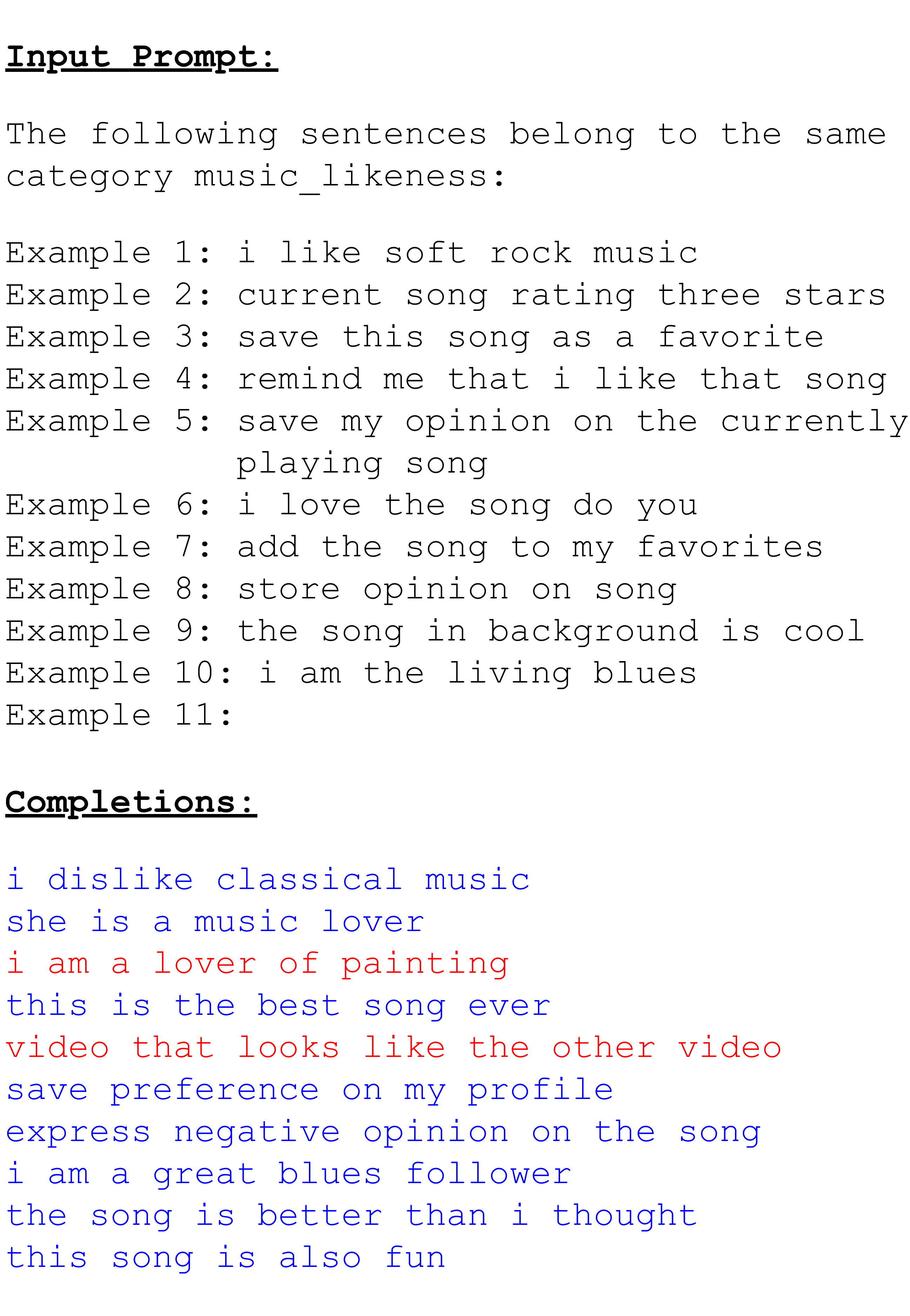}  
    \caption{\textbf{Generation Process.} Given a seed intent (here, music\_likeness) and $K$(=10) available examples for that intent, we construct a prompt following the shown template. Note that the last line of the prompt is incomplete (there is no new line character.) We then feed this prompt to a GPT-3 engine, which generates some completions of the prompt. In this example, {{\color{red} \texttt{red text}} denotes unfaithful examples and \color{blue} \texttt{blue text}} denotes faithful examples. \textbf{Note:} For brevity, we only show ten generated sentences.}
    \label{fig:method}
\end{figure}

There are two particularly promising paradigms for data augmentation in natural language processing that use pretrained language models (LMs)~\citep{peters_deep_2018,devlin_bert_2018}.
The first family of methods fine-tunes an LM on task-specific data and generates new examples using the fine-tuned LM~\citep{wu_conditional_2018,kumar_closer_2019,kumar_data_2021,anaby-tavor_not_2020,lee_neural_2021}. A limitation of these methods is that, in a real-world scenario, task-specific data is scarce   and fine-tuning an LM can quickly become the bottleneck.
The second family of methods sidesteps this bottleneck by employing off-the-shelf pretrained LMs such as GPT-3~\citep{brown_language_2020} to directly generate text without any task-specific fine-tuning. In particular, data generation by the GPT3Mix approach by ~\citet{yoo-etal-2021-gpt3mix-leveraging} boosts performance on multiple classification tasks; however, they only consider tasks with few (up to 6) classes and easy-to-grasp class boundaries (e.g., \textit{positive} and \textit{negative}).

This work studies the applicability of massive off-the-shelf LMs, such as GPT-3 and GPT-J~\cite{gpt-j} to perform effective data augmentation for intent classification (IC) tasks.
In IC, the end goal is to predict a user's intent given an utterance, i.e., what the user of a task-oriented chatbot wants to accomplish.
Data augmentation for IC is particularly challenging because the generative model must distinguish between a large number (in practice up to several hundreds) of fine-grained intents that can be semantically very close to each other. Prior methods such as GPT3Mix prompt the model with the names of all classes as well as a few examples from randomly chosen classes. We test GPT3Mix for one and observe that such approaches are poorly suitable for intent classification tasks with tens or hundreds of possible intents.
Instead, in this study, we use a simple prompt structure that focuses on a single seed intent (see Figure \ref{fig:method}) as it combines the intent's name and all available examples.


Our experiments primarily focus on few-shot IC on four prominent datasets: CLINC150~\citep{larson_evaluation_2019}, HWU64~\citep{XLiu.etal:IWSDS2019}, Banking77~\citep{casanueva-etal-2020-efficient}, and SNIPS~\citep{coucke_snips_2018}. We also consider a partial few-shot setup to compare to the Example Extrapolation (Ex2) approach by \citet{lee_neural_2021} who use a similar prompt but fine-tune the LM instead of using it as is. The main findings of our experiments are as follows: 
\textbf{(1) }GPT-generated samples boost classification accuracy when the considered intents are well-distinguished from each other (like in CLINC150, SNIPS).
\textbf{(2)} On more granular datasets (namely HWU64 and Banking77), we find that GPT struggles in distinguishing between different confounding intents.
\textbf{(3)} A small-scale study to further understand this behaviour suggests that GPT could be used as a classifier to filter out unfaithful examples and enhance the quality of the generated training set. Additionally, we investigate how valuable the generated data could be if relabelled by a human. Using an oracle model, we show that \textbf{(4)} the human labelling of GPT-generated examples can further improve the performance of intent classifiers, and that \textbf{(5)} LM-generated data has a higher relabelling potential compared to edit-based augmentation techniques, such as Easy Data Augmentation (EDA)~\citep{wei_eda_2019}.

\section{Method}
We consider training an intent classifier, where an intent is a type of request that the conversational agent supports; e.g. the user may want to change the language of the conversation, play a song, transfer money between accounts, etc. However, collecting many example utterances that express the same intent is difficult and expensive. Therefore, this paper experiments with a straightforward method to augment the training data available for an intent: creating prompts from the available examples and feeding them to a large language model such as GPT-3~\citep{brown_language_2020}.
Figure~\ref{fig:method} illustrates the process of data generation for an intent with $K$ available examples.



\section{Experimental Setup}
\subsection{Datasets}
We use four intent classification datasets in our experiments with varying levels of granularity among intents. CLINC150~\citep{larson_evaluation_2019}, HWU64~\cite{XLiu.etal:IWSDS2019} are multi-domain datasets, each covering a wide range of typical task-oriented chatbot domains, such as playing music and setting up alarms. Importantly, the CLINC150 task also contains examples of out-of-scope (OOS) utterances that do not correspond to any of CLINC's 150 intents. Banking77~\cite{casanueva-etal-2020-efficient} is a single domain dataset with very fine-grained banking-related intents. Finally, the SNIPS~\citep{coucke_snips_2018} dataset contains 7 intents typical for the smart speaker usecase. We refer the reader to Table \ref{tab:datastats} for exact statistics of all used datasets.

\begin{table}[]
    \small 
    \centering
    \begin{tabular}{ccccc}
         \toprule
                 & CLINC150 & SNIPS & HWU64 & Banking77 \\
         \midrule
         domains & 10 & 1 & 18 & 1\\
         intents & 150 & 7 & 64 & 77\\
         \multirowcell{2}{train \\ examples} & 15000 & 13084 & 8954* & 9002* \\
          & (100) & & & \\
         \multirowcell{2}{val. \\ examples} & 3000 & 700 & 1076* & 1001* \\
          & (100) & & & \\
         \multirowcell{2}{test \\ examples} & 4500 & 700 & 1076 & 3080 \\
          & (1000) & & & \\
        \bottomrule          
    \end{tabular}
    \caption{Statistics of the intent classification datasets that we use in our experiments. * indicates that we split the original data into training and validation instead of using a split provided by the dataset authors. For CLINC150, the number of out-of-scope examples in different data partitions is given in parenthesis. }
    \label{tab:datastats}
\end{table}

\subsection{Setup}
The main data-scarce setup that we consider in this work is the \textit{few-shot setup}, where only $K=10$ training examples are available for every intent of interest.  Additionally, to compare to example extrapolation with fine-tuned language models as proposed by \citet{lee_neural_2021}, we consider a \textit{partial few-shot setup}. In the latter setting, we limit the amount of training data only for a handful of \textit{few-shot intents}\footnote{We use the truncation heuristic provided by~\citet{lee_neural_2021}: \url{https://github.com/google/example_extrapolation/blob/master/preprocess_clinc150.py}} and use the full training data for others. 
When data augmentation is performed, we augment the few-shot intents to have $N$ examples, where $N$ is the median number of examples per intent of the original data. 

To precisely describe the training and test data in all settings, we will use $D_{part}$ to refer to dataset parts, i.e. train, validation, and test.
In addition, we use $D_F$ and $D_M$ to refer to data-scarce and data-rich intents (the latter only occur in the partial few-shot setting).
This notation is defined for all parts, 
therefore, $D_{part} = D_{\{F, part\}} \cup D_{\{M, part\}}$, $\forall \, part \in \{train, val, test\}$.
When GPT-3 or a baseline method is used to augment the training data we generate $N - K$ examples per intent and refer to the resulting data as $\tilde{D}_{F, train}$.  We experiment with four different-sized GPT-3  models\footnote{\url{https://beta.openai.com/docs/engines}}  by OpenAI and GPT-J by EleutherAI\footnote{\url{https://github.com/kingoflolz/mesh-transformer-jax/}} to obtain $\tilde{D}$. The four GPT-3 models are: Ada, Babbage, Curie, and Davinci. In order, Ada is the smallest model  and Davinci is the largest. Model sizes of GPT-3 engines are not known precisely but are estimated by Eleuther AI to be between ~300M and ~175B parameters\footnote{\url{https://blog.eleuther.ai/gpt3-model-sizes/}}.

\subsection{Training and Evaluation}
\label{sec:training_setup}
We fine-tune BERT-large~\citep{devlin_bert_2018} on the task of intent classification by adding a linear layer on top of the \texttt{[CLS]} token~\citep{wolf2019huggingface}.
In all setups we use the original validation set for tuning the classifier's training hyperparameters. We chose to use the full validation set as opposed to a few-shot one to avoid issues with unstable hyperparameter tuning and focus on assessing the quality of the generated data. 

\paragraph{Full few-shot.}
In this setup, we treat \textit{all} the intents as few-shot and evaluate our method on the following three scenarios: (i)~\textbf{Baseline}: all the intents are truncated to $K=10$ samples per intent, (ii)~\textbf{Augmented}: $\tilde{D}_{\{F, train\}}$ is generated using GPT and models are trained on $D_{\{F, train\}} \cup \tilde{D}_{\{F, train\}}$ and (iii)~\textbf{EDA-baseline}: same as above, but $\tilde{D}_{\{F, train\}}$ is generated using Easy Data Augmentation (EDA) by \citet{wei_eda_2019}. 
For each scenario, we report the 1) overall in-scope accuracy on the complete test set $D_{test}$, i.e. intent classification accuracy excluding OOS samples in the test set, and 2) few-shot classification accuracy of the models on $D_{\{F, test\}}$.
For CLINC150, we also report out-of-scope recall (OOS recall) on $D_{test}$ that we compute as the percentage of OOS examples that the model correctly labelled as such.

The purpose of this setting is to estimate what further gains can be achieved if the data generated by GPT were labelled by a human.
We train an oracle $\mathcal{O}$ on the full training data $D_{train}$. We also use $\mathcal{O}$ to assess the quality of the generated data. Namely, we compute \textit{fidelity} of the generated data as the ratio of generated utterances that the oracle labels as indeed belonging to the intended seed intent. A higher fidelity value means that the generated samples are more faithful to original data distribution.


\paragraph{Partial few-shot.} In this setup, we train $\mathcal{S}$ intent classifiers, choosing different \textit{few-shot intents} every time to obtain $D_F$.
We then average the metrics across these $\mathcal{S}$ runs.
For CLINC150, $\mathcal{S}=10$ corresponding to the 10 different domains, whereas for SNIPS, $\mathcal{S}=7$ corresponding to the 7 different intents.
We evaluate our method on the following three scenarios introduced by \citet{lee_neural_2021}: (i)~\textbf{Baseline}: models are trained without data augmentation on $D_{\{F, train\}} \cup D_{\{M, train\}}$. (ii)~\textbf{Upsampled}: $D_{\{F, train\}}$ is upsampled to have $N$ examples per intent.  Then models are trained on upsampled $D_{\{F, train\}} \cup D_{\{M, train\}}$. (iii)~\textbf{Augmented}: models are trained on $D_{\{F, train\}}
\cup \tilde{D}_{\{F, train\}} \cup D_{\{M, train\}}$. For each scenario in this setup, we report the overall in-scope classification accuracy (and OOS Recall for CLINC150).

 
For both partial few-shot and full few-shot settings, we report means and standard deviations over 10 repetitions of each experiment.

\section{Experimental Results}
\label{sec:results}

\begin{table*}[t!]
    \centering
    {%
    \begin{tabular}{lccccc}
        \toprule
         & \multicolumn{2}{c}{\textbf{CLINC150}} & \multicolumn{1}{c}{\textbf{HWU64}} &
         \multicolumn{1}{c}{\textbf{Banking77}} & \multicolumn{1}{c}{\textbf{SNIPS}} \\
         \cmidrule(r){2-6}
         \textbf{Model}  & \textbf{IA (96.93)} & \textbf{OR (42.9)} & \textbf{IA (92.75)} & \textbf{IA (93.73)} & \textbf{IA (98.57)} \\
        \midrule
        EDA & 92.66 (0.40) & 43.81 (2.03) & 83.67 (0.48) & 83.96 (0.66) & 92.50 (1.61) \\
        Baseline (Ours) & 90.28(0.49) & 50.18(1.14) & 81.43 (0.57) & 83.35 (0.59) & 89.69 (1.63) \\

        \midrule
        \multicolumn{6}{c}{\textbf{Augmented}} \\
        \midrule
         Ada (Ours) & 91.31 (0.34) & 21.69 (1.57) & 79.68 (0.83) & 79.30 (0.42) & 94.27 (0.52) \\
         Babbage (Ours) & 92.72 (0.33) & 22.99 (2.39) & 81.86 (0.78) & 80.31 (0.41) & 94.74 (0.67) \\
         Curie (Ours) & 93.37 (0.21) & 25.85 (1.49) & 82.85 (0.70) & 83.50 (0.44) & 94.73 (0.62) \\
         GPT-J (Ours) & 93.25 (0.19) & 24.02 (1.45) & 81.78 (0.56) & 82.32 (0.90) & 95.19 (0.61) \\
         Davinci (Ours) & 94.07 (0.18) & 27.36 (1.08) & 82.79 (0.93) & 83.60 (0.45) & 95.77 (0.86) \\
        \midrule
        \multicolumn{6}{c}{\textbf{Augmented + Relabelled}} \\
        \midrule
        EDA & 93.43 (0.22) & 48.56 (1.84) & 85.58 (0.73) & 84.82 (0.57) & 94.91 (0.66) \\
        Ada (Ours) & 95.09 (0.16) & 41.38 (1.77) & 88.53 (0.61) & 88.45 (0.19) & 97.03 (0.18) \\
        Babbage (Ours) & 95.39 (0.17) & 40.58 (1.63) & 89.49 (0.32) & 88.86 (0.26) & 96.89 (0.49) \\
        Curie (Ours) & 95.08 (0.19) & 40.09 (2.38) & 89.78 (0.47) & 88.30 (4.64) & 96.86 (0.31) \\
        GPT-J (Ours) & 95.11 (0.13) & 43.94 (1.76) & 89.52 (0.54) & 88.94 (0.40) & 97.33 (0.38) \\
        Davinci (Ours) & 95.08 (0.13) & 40.76 (1.37) & 89.53 (0.45) & 88.89 (0.31) & 97.03 (0.38) \\
        \bottomrule
    \end{tabular}
    }
    \caption{\textbf{Full few-shot results on CLINC150, HWU64, Banking77, and SNIPS datasets.} \textbf{IA}: Inscope Accuracy (mean (std)). \textbf{OR}: OOS-Recall (mean (std)). Towards the top of the table, we also report the test set performance (enclosed in parentheses) when all examples are used for fine-tuning (without any augmentation.)}
    \label{tab:Few-Shot-Results}
\end{table*}

\begin{table*}[t!]
    \centering
    \resizebox{\textwidth}{!}
    {%
    \begin{tabular}{lcccccc}
        \toprule
          & & \multicolumn{3}{c}{\textbf{CLINC150}} & \multicolumn{2}{c}{\textbf{SNIPS}}\\
         \cmidrule(r){3-5}\cmidrule(r){6-7}
          & &\multicolumn{2}{c}{\textbf{Overall}} & \multicolumn{1}{c}{\textbf{Few-shot}} & \multicolumn{1}{c}{\textbf{Overall}} & \multicolumn{1}{c}{\textbf{Few-shot}}\\ \cmidrule(r){3-4}\cmidrule(r){5-5}\cmidrule(r){6-6}\cmidrule(r){7-7}
                 & \textbf{Classifier}  & \textbf{IA} & \textbf{OR} & \textbf{A} & \textbf{IA} & \textbf{A} \\
        \midrule
        Baseline$^\spadesuit$ & T5 & 97.4 & - & 93.7 & 95.2 & 74.0 \\
        Upsampled$^\spadesuit$ & T5 &  97.4 & - & 94.4 & 95.9 & 80.0 \\
        Augmented (Ex2)$^\spadesuit$ & T5 &  97.4 & - & 95.6 & 97.8 & 94.0 \\
        \midrule
        Baseline (ours) &  BERT & 96.28 (0.06) & 39.14 (0.82) & 91.36 (0.47) & 95.47 (0.45) & 78.38 (3.34) \\
        Upsample (ours) &  BERT & 96.20 (0.05) & 40.21 (0.59) & 90.93 (0.19) & 95.29 (0.37) & 79.28 (2.05) \\
        \midrule
        Augmented (Ada) & BERT & 96.16 (0.05) & 34.37 (0.27) & 92.60 (0.15) & 97.30 (0.24) & 94.41 (0.72) \\
        Augmented (Babbage) & BERT & 96.39 (0.06) & 35.71 (0.46) & 93.66 (0.21) & 97.46 (0.25) & 95.31 (0.74) \\
        Augmented (Curie) & BERT & 96.41 (0.06) & 36.77 (0.93) & 93.90 (0.21) & 97.37 (0.19) & 94.79 (0.64) \\
        Augmented (GPT-J) & BERT & 96.38 (0.05) & 35.91 (0.94) & 93.85 (0.25) & 97.59 (0.21) & 96.08 (0.39) \\
        Augmented (Davinci) & BERT & 96.45 (0.03) & 37.52 (0.54) & 94.28 (0.24) & 97.66 (0.21) & 96.52 (0.35) \\
        \bottomrule
    \end{tabular}
    }
    \caption{\textbf{Partial few-shot results on CLINC150 and SNIPS datasets.} Refer to Section~\ref{sec:training_setup} for more details. \textbf{IA}: Inscope accuracy (mean (std)). \textbf{OR}: OOS Recall (mean (std)). \textbf{A}: Accuracy (mean (std)). $^\spadesuit$~\cite{lee_neural_2021}.}
    \label{tab:Partial-Few-Shot-Results}
\end{table*}

\begin{figure}[t!]
\centering
\begin{subfigure}{\linewidth}
    \includegraphics[width=\textwidth]{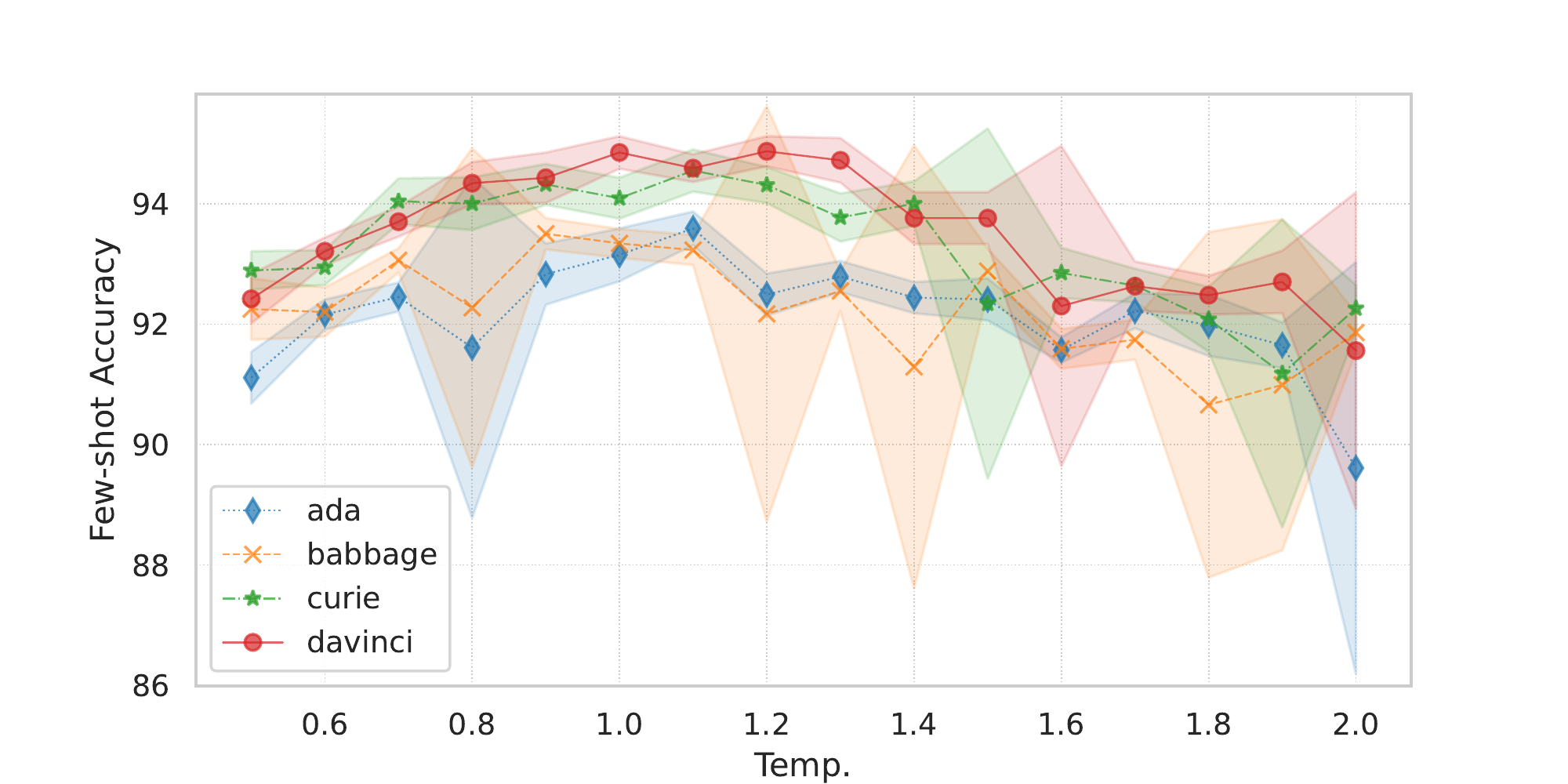}
    \caption{Temperature v/s Few-shot accuracy}
    \label{fig:fs_acc}
\end{subfigure}
\hfill
\begin{subfigure}{\linewidth}
    \includegraphics[width=\textwidth]{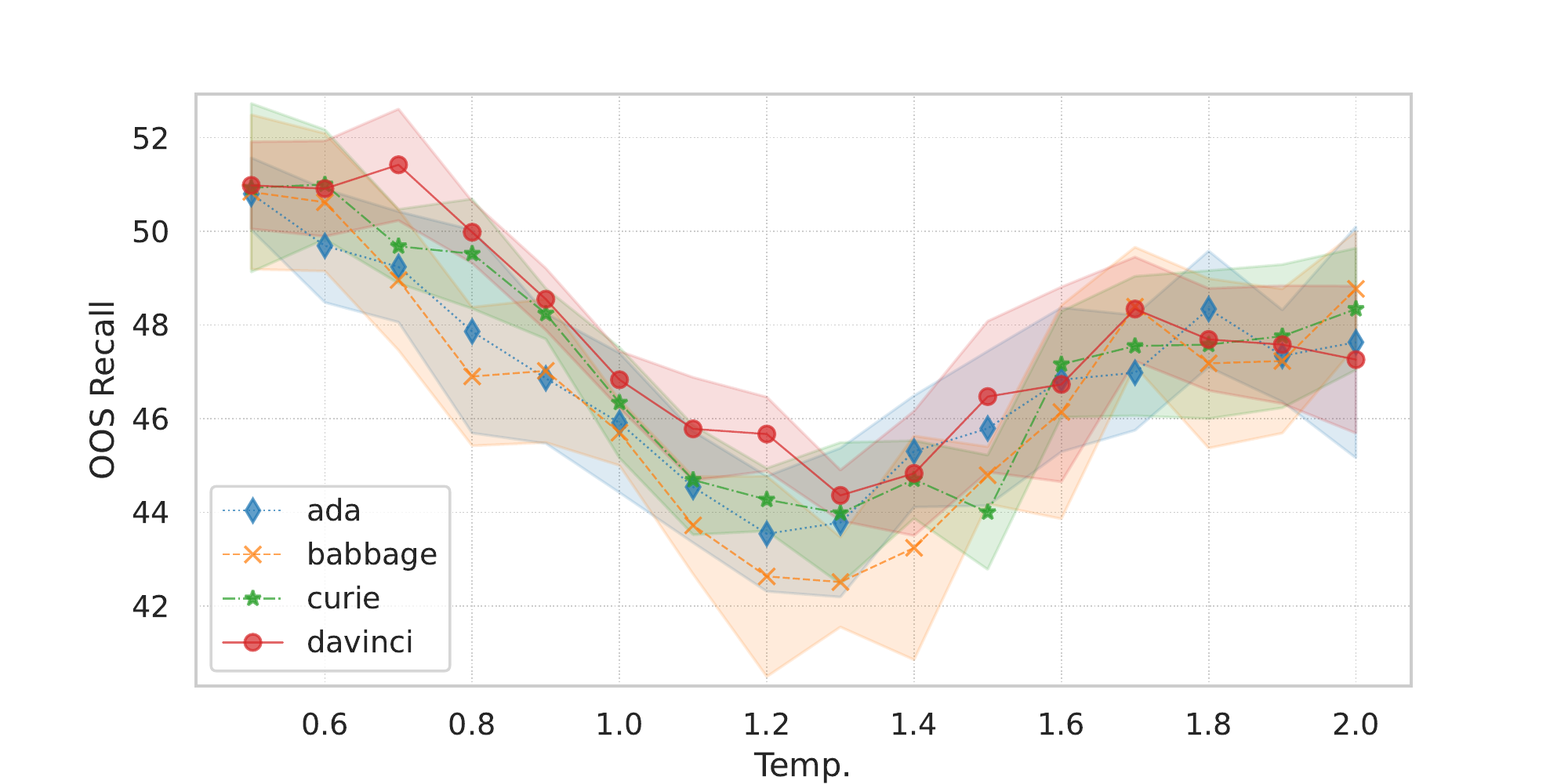}
    \caption{Temperature v/s OOS recall}
    \label{fig:or}
\end{subfigure}
\hfill
\begin{subfigure}{\linewidth}
    \includegraphics[width=\textwidth]{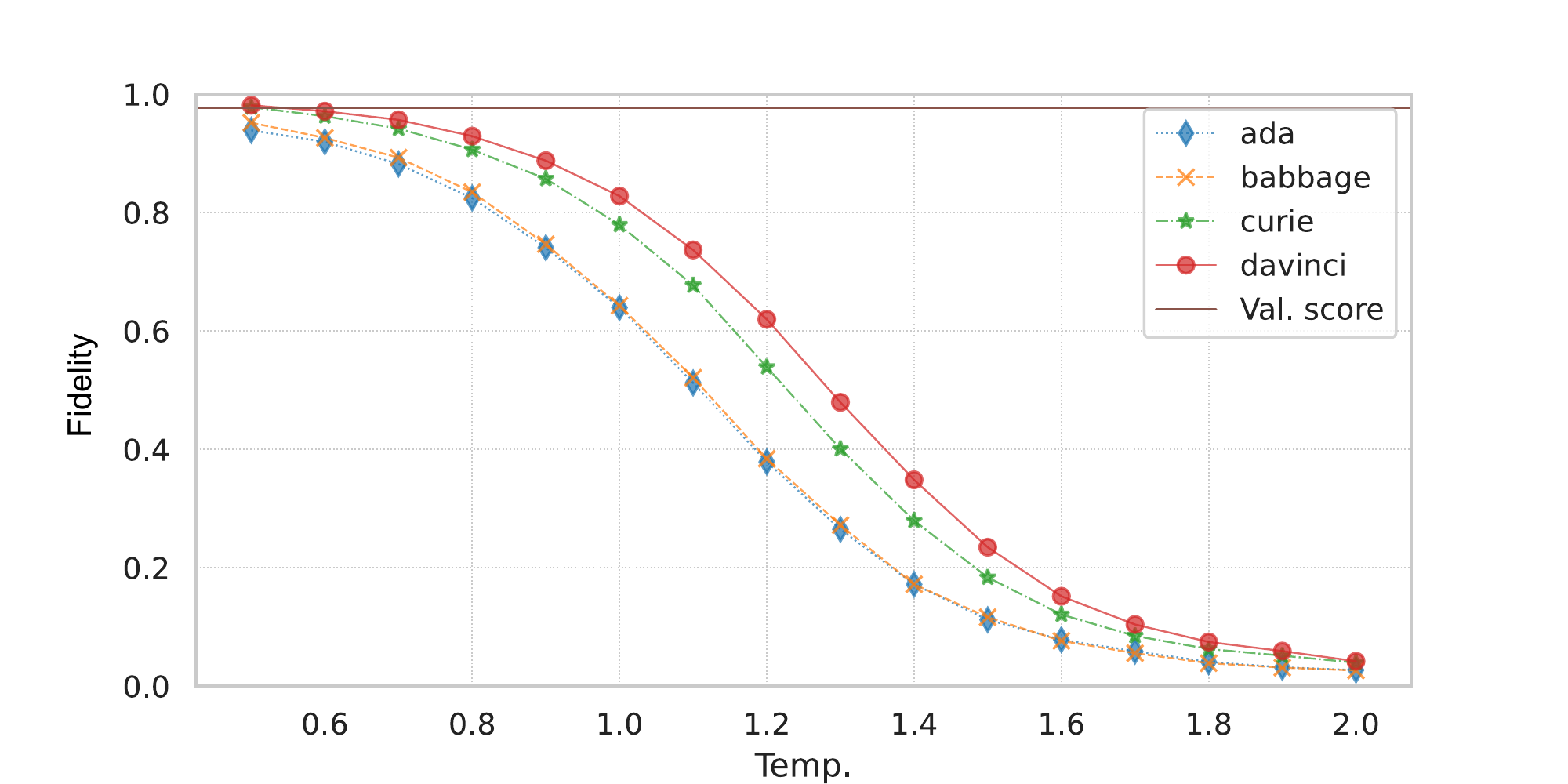}
    \caption{Temperature v/s Fidelity}
    \label{fig:fidelity}
\end{subfigure}
        
\caption{\textbf{Partial few-shot validation performance for different GPT-3 models and temperatures.} (a) few-shot accuracy, (b) OOS recall of intent classifiers trained on augmented sets, and (c) fidelity measured as the accuracy of the oracle on the augmented sets.}
\label{fig:temperatures}
\end{figure}

\begin{figure}[t!]
\centering
\begin{subfigure}{\linewidth}
    \includegraphics[width=\textwidth]{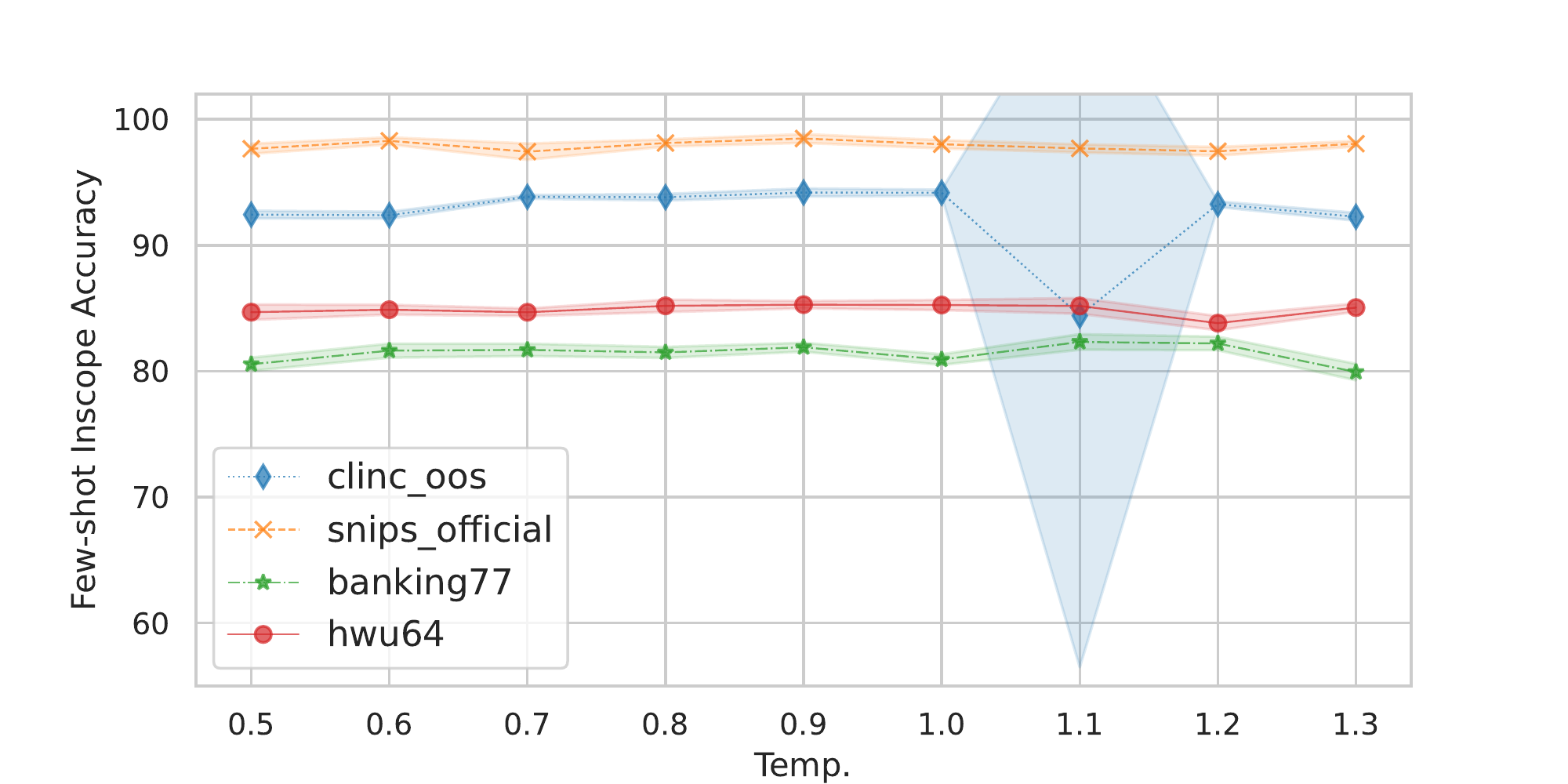}
    \caption{Temperature v/s Few-shot inscope accuracy}
    \label{fig:fs_acc_ds}
\end{subfigure}
\hfill
\begin{subfigure}{\linewidth}
    \includegraphics[width=\textwidth]{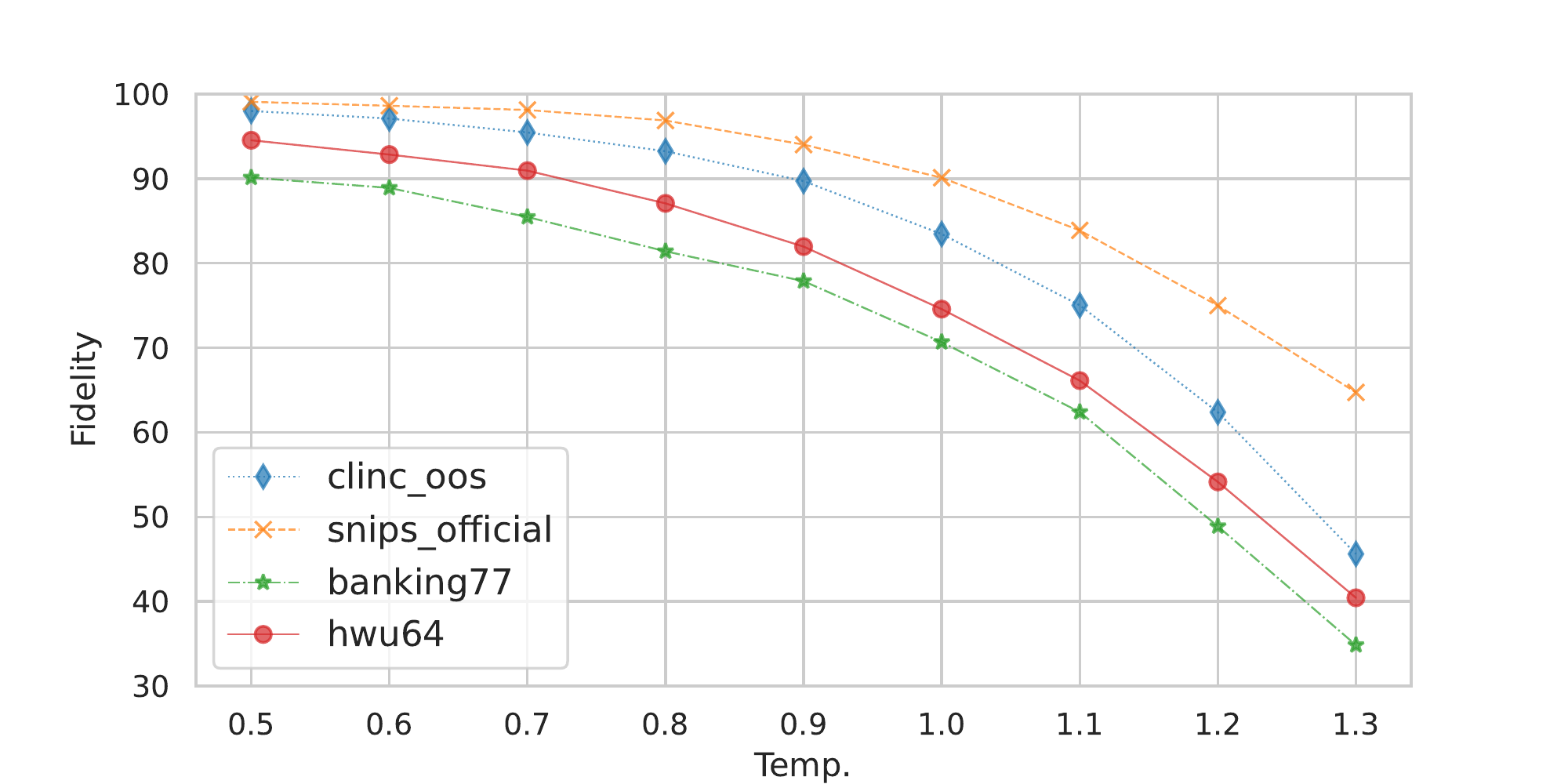}
    \caption{Temperature v/s Fidelity}
    \label{fig:fidelity_ds}
\end{subfigure}
        
\caption{\textbf{Full few-shot validation performance for different GPT-J temperatures on different datasets.} (a) few-shot inscope accuracy of intent classifiers trained on augmented sets, and (b) fidelity (oracle accuracy) of augmented sets generated by GPT-J with different temperatures.}
\label{fig:fidelities}
\end{figure}

\paragraph{Full few-shot.} Table~\ref{tab:Few-Shot-Results} shows the results of our few-shot experiments. For CLINC150 and SNIPS, data augmentation with GPT-3 is very effective as it leads to respective accuracy improvements of up to approximately 3.7\% and 6\% on these tasks. These improvements are larger than what the baseline EDA method brings, namely 2.4\% and 2.9\% for CLINC150 and SNIPS. Importantly, using larger GPT models for data augmentation brings significantly bigger gains. Data augmentation results on Banking77 and HWU64 are, however, much worse, with no or little improvement upon the plain few-shot baseline. 
We present a thorough investigation of this behaviour in Section~\ref{sec:analysis}. One can also see that data augmentation with GPT models lowers the OOS recall.

Next, we observe that relabelling EDA and GPT-generated sentences by the oracle gives a significant boost to accuracies across the board, confirming our hypothesis that human inspection of generated data could be fruitful.
Importantly, we note that the magnitude of improvement for EDA is less than for GPT models.
This suggests that GPT models generate more diverse data that can eventually be more useful after careful human inspection.
Lastly, relabelling also improves OOS recall on CLINC150, which is due to the fact that much of the generated data was labelled as OOS by the oracle.

\paragraph{Partial few-shot.} Table~\ref{tab:Partial-Few-Shot-Results} shows the results of our partial few-shot experiments on CLINC150 and SNIPS.
By augmenting the dataset with GPT-generated samples, the few-shot accuracy improves by up to 2.92\% on CLINC150 and 18.14\% on SNIPS compared to the baseline setting. 
Our method achieves competitive results compared to Ex2~ \citep{lee_neural_2021}, both in terms of absolute accuracies and the relative gains brought by data augmentation.  Note that Ex2 uses T5-XL~\cite{roberts-etal-2020-much} with nearly 3 billion parameters as its base intent classifier, while our method uses BERT-large with only 340 million parameters.

\begin{table*}[t!]
    \small
    \centering
    \resizebox{0.9\textwidth}{!}
    {%
    \begin{tabular}{|p{8cm}|c|c|}
        \hline
        \textbf{Davinci generated sentences} & \textbf{Seed Intent} & \textbf{Oracle Prediction} \\
        \hline
        \multicolumn{3}{|c|}{\textbf{HWU64}} \\
        \hline
        \colorbox{yellow}{play a song} with the word honey & music\_likeness & play\_music \\
        \hline
        you are \colorbox{yellow}{playing music} & music\_likeness & play\_music \\
        \hline
        \colorbox{yellow}{`let me hear} some of that jazz!' & music\_likeness & play\_music \\
        \hline
        i \colorbox{yellow}{really like} myspace music & play\_music & music\_likeness \\
        \hline
        \colorbox{yellow}{i love} the start lucky country music & play\_music & music\_likeness \\
        \hline
        \colorbox{yellow}{thank you} for the music & play\_music & music\_likeness \\
        \hline
        please \colorbox{yellow}{play the next song} & music\_settings & play\_music \\
        \hline
        \colorbox{yellow}{play background music} & music\_settings & play\_music \\
        \hline
        \colorbox{yellow}{play the hour long loop} of rock song & music\_settings & play\_music \\
        \hline
        need you to play that song \colorbox{yellow}{one more time} & play\_music & music\_settings \\
        \hline
        \colorbox{yellow}{skip} that song, its turkish & play\_music & music\_settings \\
        \hline
        \colorbox{yellow}{pickup the beat} or a temp track or audio plugin & play\_music & music\_settings \\
        \hline
        \multicolumn{3}{|c|}{\textbf{Banking77}} \\
        \hline
        My last attempt to top up \colorbox{yellow}{didn't} seem to \colorbox{yellow}{work}, any success? & topping\_up\_by\_card & top\_up\_failed \\
        \hline
        I tried to top off my wallet using my card but it says \colorbox{yellow}{``top up failed".} & topping\_up\_by\_card & top\_up\_failed \\
        \hline
        \colorbox{yellow}{I cannot top-up} by my cellular phone number? How do I do that? & topping\_up\_by\_card & top\_up\_failed \\
        \hline
        Can you transfer money to my Ola prepaid option? Or help me \colorbox{yellow}{top up my card to money}. They never accept my card so I always have to suffer & top\_up\_failed & topping\_up\_by\_card \\
        \hline
        Hi my app is activated on activate.co.in, but unable to top up my phone. \colorbox{yellow}{I tried credit card, debit card} and Paytm \colorbox{yellow}{but fails} & top\_up\_failed & topping\_up\_by\_card \\
        \hline
        I try to top up my card but it's not going through. It's \colorbox{yellow}{still on pending status.} Do I need to wait or did I do something wrong & top\_up\_failed & pending\_top\_up \\
        \hline
        I tried top-up with my card but notification \colorbox{yellow}{shows that `Pending'.} This has been happening since last night. Can you tell me what's going on & top\_up\_failed & pending\_top\_up \\
        \hline
        Top up  \colorbox{yellow}{didn't go through.} & pending\_top\_up & top\_up\_failed \\
        \hline
        Did my master card \colorbox{yellow}{top-up fail?} & pending\_top\_up & top\_up\_failed \\
        \hline
    \end{tabular}%
    }
    \caption{\textbf{Davinci-generated sentences for closely-related intents in HWU64 and Banking77 datasets.} Highlighted sub-strings indicate a difference with respect to the seed intent.}
    \label{tab:gpt_confusion}
\end{table*}

\begin{table}[t!]
    {%
    \begin{tabular}{lcc}
        \toprule
        \textbf{Fidelity (3 intents) } & \textbf{HWU64} & \textbf{Banking77} \\
        \midrule
         w/o filtering (468) & 60.26 & 57.31 \\
         w/ filtering (371) & 72.51 & 65.54 \\
        \midrule
        \textbf{3-way accuracy} & & \\ 
        \midrule
        Davinci & 86.36 & 78.75 \\
        10-shot BERT-large & 82.95 & 65.54 \\        
        Full data BERT-large & 94.32 & 95.00 \\
        \bottomrule
    \end{tabular}
    }
    \caption{The impact and the accuracy of using GPT-3 as a 3-way classifier on close intent triplets from  HWU64 and Banking77 datasets. For fidelity, generated examples are rejected if the GPT-3 classifier labels them as not belonging to the seed intent. Classification accuracies are reported on the reduced validation+test sets where we only consider examples from the three confounding intents.}
    \label{tab:Results2}
\end{table}

\begin{figure}[t!]
\centering
\includegraphics[width=\linewidth]{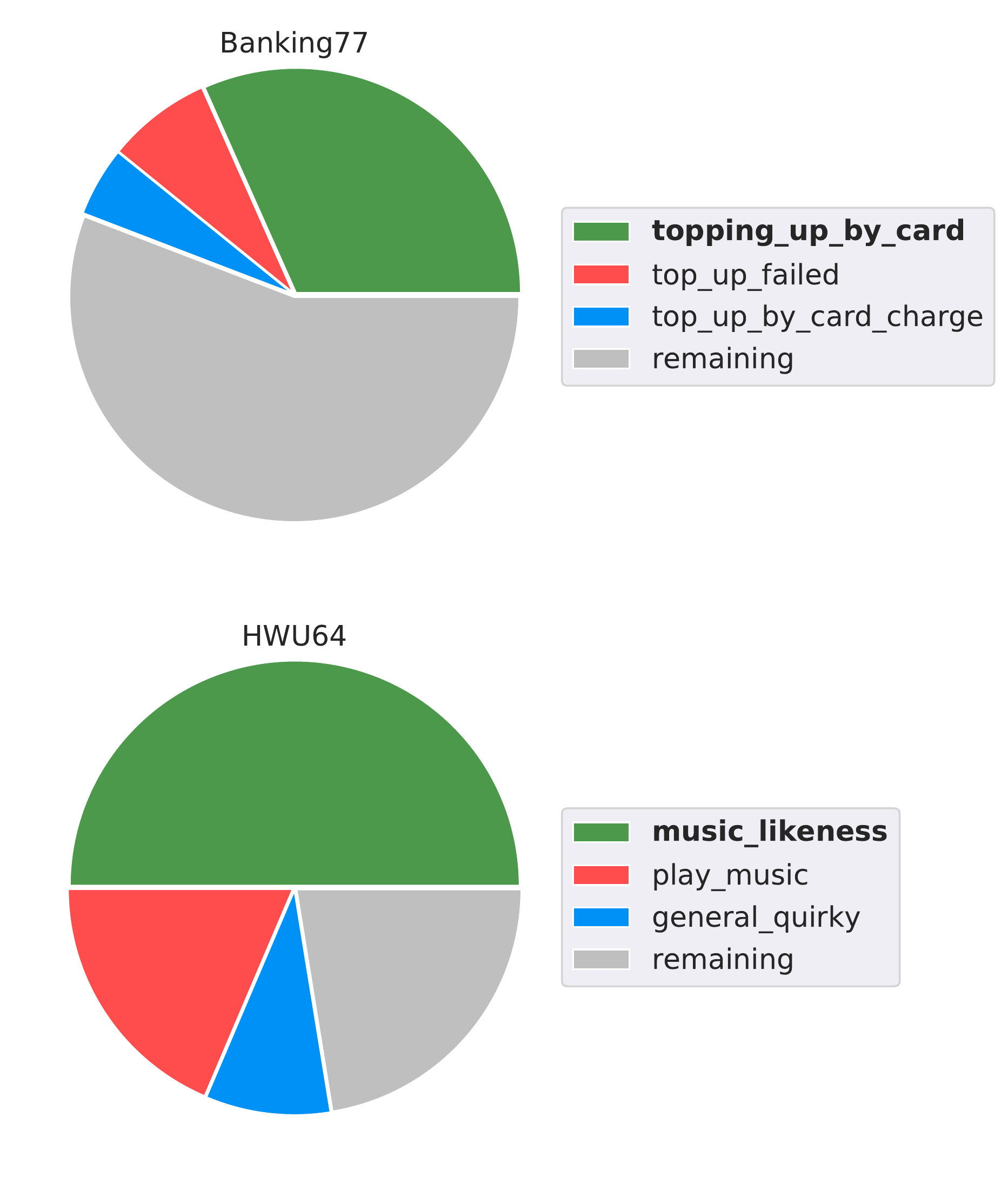}
\caption{Distribution of labels as predicted by the oracle for lowest-fidelity intents in Banking77 and HWU64 datasets (``topping\_up\_by\_card" and ``music\_likeness," respectively). Green areas denote the portion of generated sentences deemed fit by the oracle for the lowest-fidelity intents in the two datasets. Red and Blue areas respectively correspond to the most common and the second most common alternative intent predicted by the oracle.}
\label{fig:gpt_confusion}
\end{figure}

\subsection{Analysis}
\label{sec:analysis}
\textbf{Effect of GPT sampling temperature.} We investigate the impact of generation temperature on the quality and fidelity of generated data. We perform this investigation on the CLINC150 dataset using the partial few-shot setup. Results in Figure~\ref{fig:temperatures} show that, for all engines, the generated data leads to the highest classification accuracy when the generation temperature is around 1.0, although lower temperatures result in higher OOS recall. We also observe that the fidelity of the generated samples decreases as we increase the temperature (i.e. higher diversity, see Figure \ref{fig:fidelity}).
This suggests that higher fidelity does not always imply better quality samples as the language model may simply copy or produce less diverse utterances at lower temperatures. In Appendix~\ref{section:appendix}, we perform a human evaluation, reaching similar conclusions as when using an oracle to approximate fidelity. 

\noindent\textbf{Fidelity on different datasets.}
Our results in Section \ref{sec:results} show that data augmentation gains are much higher on CLINC150 and SNIPS
than on HWU64 and Banking77. To contextualize these results, we report the fidelity of GPT-J-generated data for all these tasks in Figure \ref{fig:fidelity_ds}. Across all generation temperatures, the fidelity of the generated data is higher for CLINC150 and SNIPS than for HWU64 and Banking77. For all datasets, the fidelity is higher when the generation temperature is lower; however, Figure \ref{fig:fs_acc_ds} shows that low-temperature data also does improve the model's performance.

\paragraph{Data generation for close intents.} To better understand the lower fidelity and accuracy on HWU64 and Banking77 datasets, we focus on intents with the lowest fidelities. Here, by intent fidelity, we mean the percentage of the intent's generated data that the oracle classified as indeed belonging to the seed intent. 
In the Banking77 dataset, the lowest-fidelity intent is ``topping\_up\_by\_card.'' For this intent, only 33\% of the Davinci-generated sentences were labelled as ``topping\_up\_by\_card'' by the oracle, implying that two-thirds of the sentences did not fit that intent, ``top\_up\_failed" and ``top\_up\_card\_charge" being the two most common alternatives chosen by the oracle.
Similarly, only 50\% of the Davinci-generated sentences abide by the lowest-fidelity ``music\_likeness" intent in the HWU64 dataset, ``play\_music'' and ``general\_quirky" being the most common intents among the ``unfaithful" sentences.
Figure~\ref{fig:gpt_confusion} visualizes this high percentage of unfaithful generated sentences.
It also shows the proportion of the two most common alternatives that the oracle preferred over the seed intent.
Table~\ref{tab:gpt_confusion} presents generated sentences for confounding intents in the HWU64 and Banking77 datasets.
There are clear indications of mix-up of intents, e.g., Davinci generates, ``play a song with the word honey," which should belong to ``play\_music" rather than ``music\_likeness."
There are also instances where the LM mixes two intents; for instance, Davinci generates ``Hi my app is activated on activate.co.in, but unable to top up my phone. I tried credit card, debit card and Paytm but fails," which could belong to either ``topping\_up\_by\_card" intent (as it mentions about using credit card in the context of a top up) or ``top\_up\_failed" (as the top up ultimately fails).

\subsection{Can GPT Models Understand Close Intents?}
\label{sec:close_intents} 
\begin{figure}[t!]
    \centering
    \includegraphics[width=\linewidth]{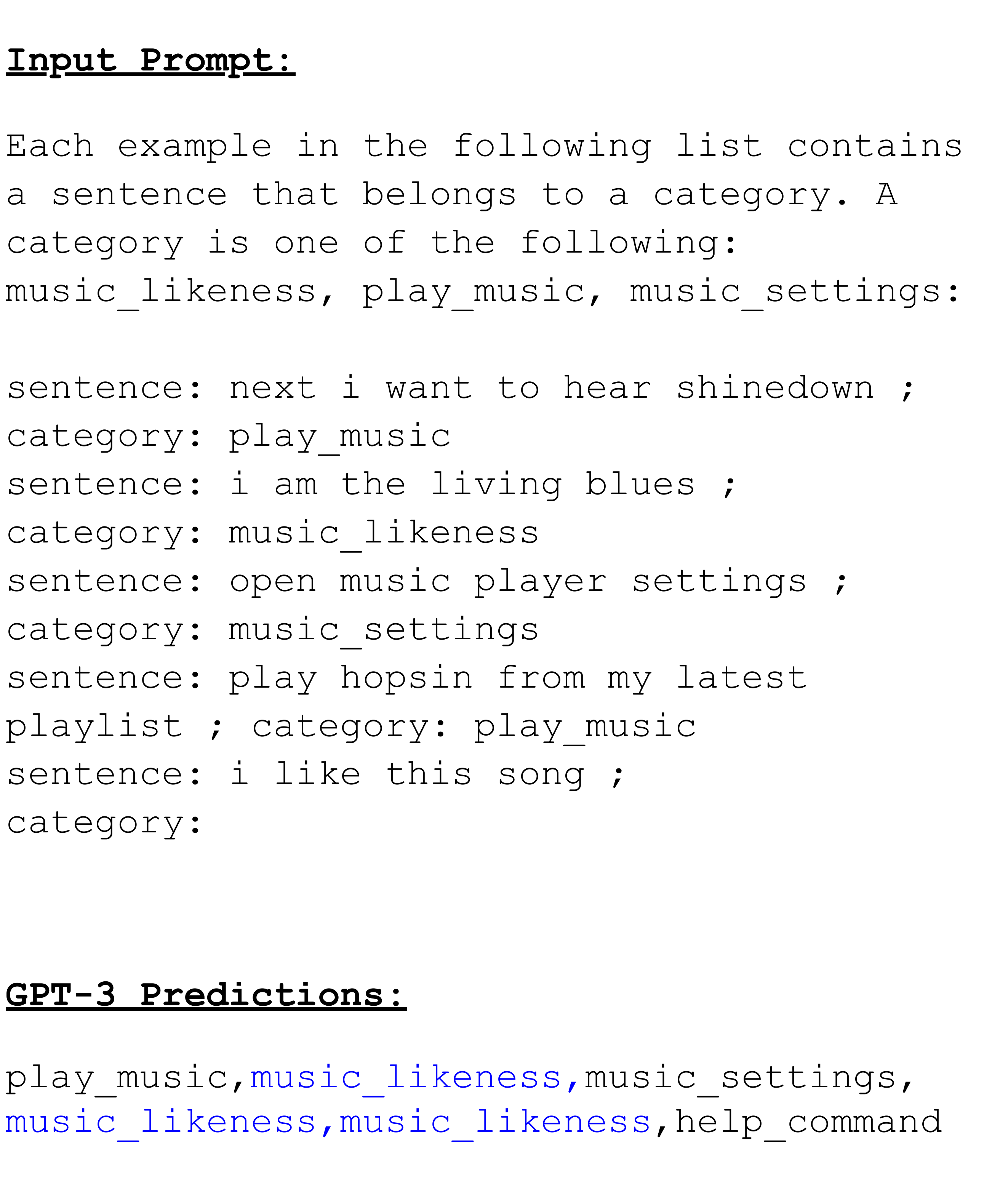}  
    \caption{\textbf{Using GPT-3 as a classifier.} Given a triplet of close intents, we mix and shuffle the multiple seed examples available for each of them. Then, we append an incomplete line to the prompt with just the generated sentence and feed it to GPT-3 multiple times.
    Among the responses, we choose the most generated in-triplet intent as the predicted intent (``music\_likeness" in the above example). \textbf{Note:} For brevity, we don't show all the seed examples and predictions.}
    \label{fig:classifier_method}
\end{figure}

We perform extra investigations to better understand
what limits GPT-3's ability to generate data accurately. We hypothesize that one limiting factor can be GPT-3's inability to understand fine-grained differences in the meanings of utterances. To verify this hypothesis, we evaluate how accurate GPT-3 is at classifying given utterances as opposed to generating new ones. Due to the limited prompt size of 2048 tokens, we can not prompt GPT-3 to predict all the intents in the considered datasets. We thus focus on the close intent triplets from HWU64 and Banking77 datasets that we use in Table~\ref{tab:gpt_confusion}. We compare the 3-way accuracy of a prompted GPT-3 classifier to the similarly-measured 3-way performance of conventional BERT-large classifiers. We prompt GPT-3 with 10 examples per intent (see Figure~\ref{fig:classifier_method}). For comparison, we train BERT-large classifiers on either the same 10 examples or the full training set. Table~\ref{tab:Results2} shows that the Davinci version of GPT-3 performs in between the 10-shot and the full-data conventional classifiers. This suggests that while GPT-3's understanding of nuanced intent differences is imperfect, it could still be sufficient to improve the performance of the downstream few-shot model. Inspired by this finding, we experiment with using GPT-3's classification abilities to improve the quality of generated data. Namely, we reject the generated utterances that GPT-3 classifies as not belonging to the seed intent. For both HWU64 and Banking77, this filtering method significantly improves the fidelity of the generated data for the chosen close intent triplets.

\subsection{Comparison with GPT3Mix}
\label{gpt3mix_results}
To test our initial hypothesis that prior methods such as GPT3Mix are not suitable for intent classification, we experiment with the said method on the CLINC150 dataset using Curie. Specifically, we include an enumeration of the 150 intent names in the prompt and randomly select one example for $K$ intents.
We observe a poor in-scope accuracy of 86.33\% in the \textit{Augmented} scenario\footnote{Average of 10 runs with a standard deviation of 1.17}.
Furthermore, the generated samples have low fidelity (27.96\%).
We also test a mixture of GPT3Mix prompt and our prompt where we include all the $K$ examples for the seed intent instead of 1 example per $K$ randomly sampled intents.
This mixed variant also performs poorly on CLINC150 and only achieves an in-scope accuracy of 86.05\%\footnote{Average of 10 runs with a standard deviation of 0.59} and a fidelity of 33.56\%.
Our interpretation of this result is that GPT cannot handle the long list of 150 intent names in the prompt.

\section{Related Work}
The natural language processing literature features diverse data augmentation methods. Edit-based methods such as Easy Data Augmentation apply rule-based changes to the original utterances  to produce new ones \citep{wei_eda_2019}. In back-translation methods \citep{sennrich_improving_2016} available examples are translated to another language and back. Recently, data augmentation with fine-tuned LMs has become the dominant paradigm \citep{wu_conditional_2018,kumar_closer_2019,kumar_data_2021,anaby-tavor_not_2020,lee_neural_2021}. Our simpler method sidesteps LM-fine-tuning and directly uses off-she-shelf LMs as is.

The data augmentation approach that is closest to the one we use here is GPT3Mix by \citet{yoo-etal-2021-gpt3mix-leveraging}. A key part of the GPT3Mix prompt is a list of names of all possible classes (e.g. ``The sentiment is one of `positive' or `negative''').
The LM is then expected to pick a random class from the list and generate a new example as well as the corresponding label.
However, this approach does not scale to intent classification setups, which often feature hundreds of intents (see Section~\ref{gpt3mix_results}).
Therefore, we choose a different prompt that encourages the model to extrapolate between examples of a seed intent similarly to \citep{lee_neural_2021}.

Other work on few-shot intent classification explores fine-tuning dialogue-specific LMs as classifiers as well as using similarity-based classifiers instead of MLP-based ones on top of BERT \citep{vulic-etal-2021-convfit}. We believe that improvements brought by data augmentation would be complementary to the gains brought by these methods.

Lastly, our method to filter out unfaithful GPT generations is related to the recent work by~\citet{wang_want_2021} that proposes using GPT3 for data labelling. A crucial difference with respect to our work, however, is that we use GPT-3 for rejecting mislabelled samples rather than proposing labels for unlabelled samples.

\section{Conclusion}
We propose a prompt-based method to generate intent classification data with large pretrained language models. Our experiments show that generated data can be helpful as additional labelled data for some tasks, whereas, for other tasks, the generated data needs to be either relabelled or filtered to be helpful. We show that a filtering method that recasts the same GPT model as a classifier can be effective. Our filtering method, however, requires knowing the other intents that the generated data is likely to belong to instead of the seed intent. Future work can experiment with heuristics for approximately identifying the most likely actual intents for the generated utterances. This would complete a data generation and filtering pipeline that, according to our preliminary results in Section~\ref{sec:close_intents} here, could be effective. Other filtering methods could also be applied, such as looking at the likelihood of the generated utterances as explored in a concurrent work by~\citet{meng2022generating}. Lastly, an interesting future work direction is identifying which generated utterances most likely need a human inspection. 

\section{Ethical Considerations}
As discussed for the GPT3Mix method in~\citet{yoo-etal-2021-gpt3mix-leveraging}, using large language models for data augmentation presents several challenges:
they exhibit social biases and are prone to generating toxic content.
Therefore, samples generated using our prompting-based approach need to be considered carefully.

To address such ethical concerns, human inspection would be the most reliable way to identify and filter out problematic generations. The practitoners who apply our method may also consider debiasing the language model before using it for generation~\citep{schick_exploiting_2021}.


\bibliography{dima,common}
\bibliographystyle{acl_natbib}

\clearpage
\appendix
\section*{Appendix}
\section{Human Evaluation}
\label{section:appendix}
In Figure~\ref{fig:temperatures} we evaluate the fidelity of the samples generated by GPT-3 with respect to the original set of sentences used to prompt it. Fidelity is approximated by the classification performance of an "oracle" intent classifier trained on the whole dataset ($D_{train} \cup D_{test}$) and evaluated over the generated samples. In order assess whether the oracle predictions are comparable to those of a human, we perform a human evaluation study.

\begin{figure}[h]
\includegraphics[width=\linewidth]{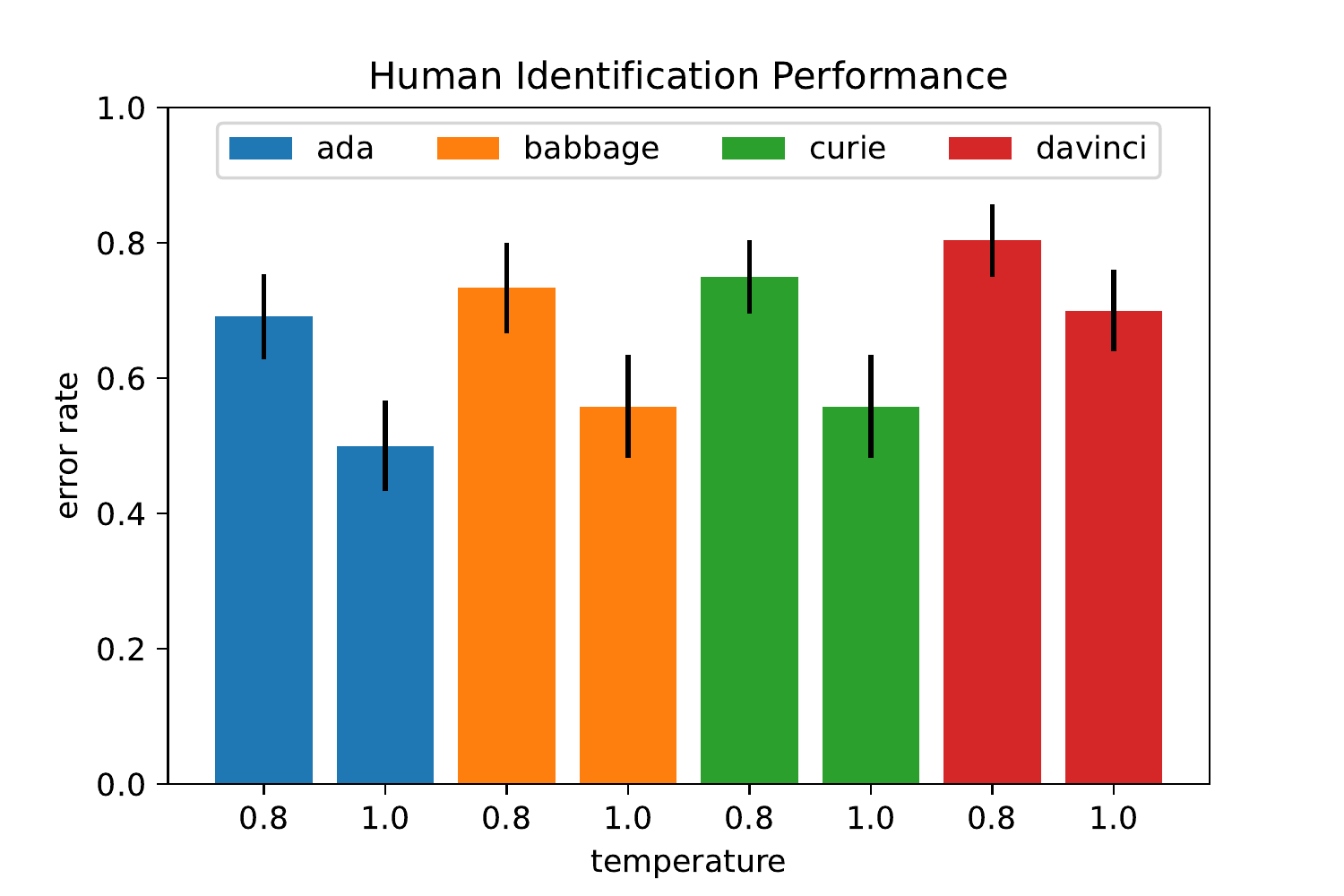}
\caption{\textbf{Human evaluation.} Error rate of human evaluators at the task of finding whether any sentence in a group of 5 was generated by GPT-3 or not. Each color represents a different GPT-3 engine. Higher error rate indicates that humans could not correctly identify generated samples and thus it also indicates higher fidelity. The standard error is displayed as a vertical line on top of each bar.}
\label{fig:human_evaluation}
\end{figure}
\begin{figure}[h]
    \centering
    \includegraphics[width=\linewidth]{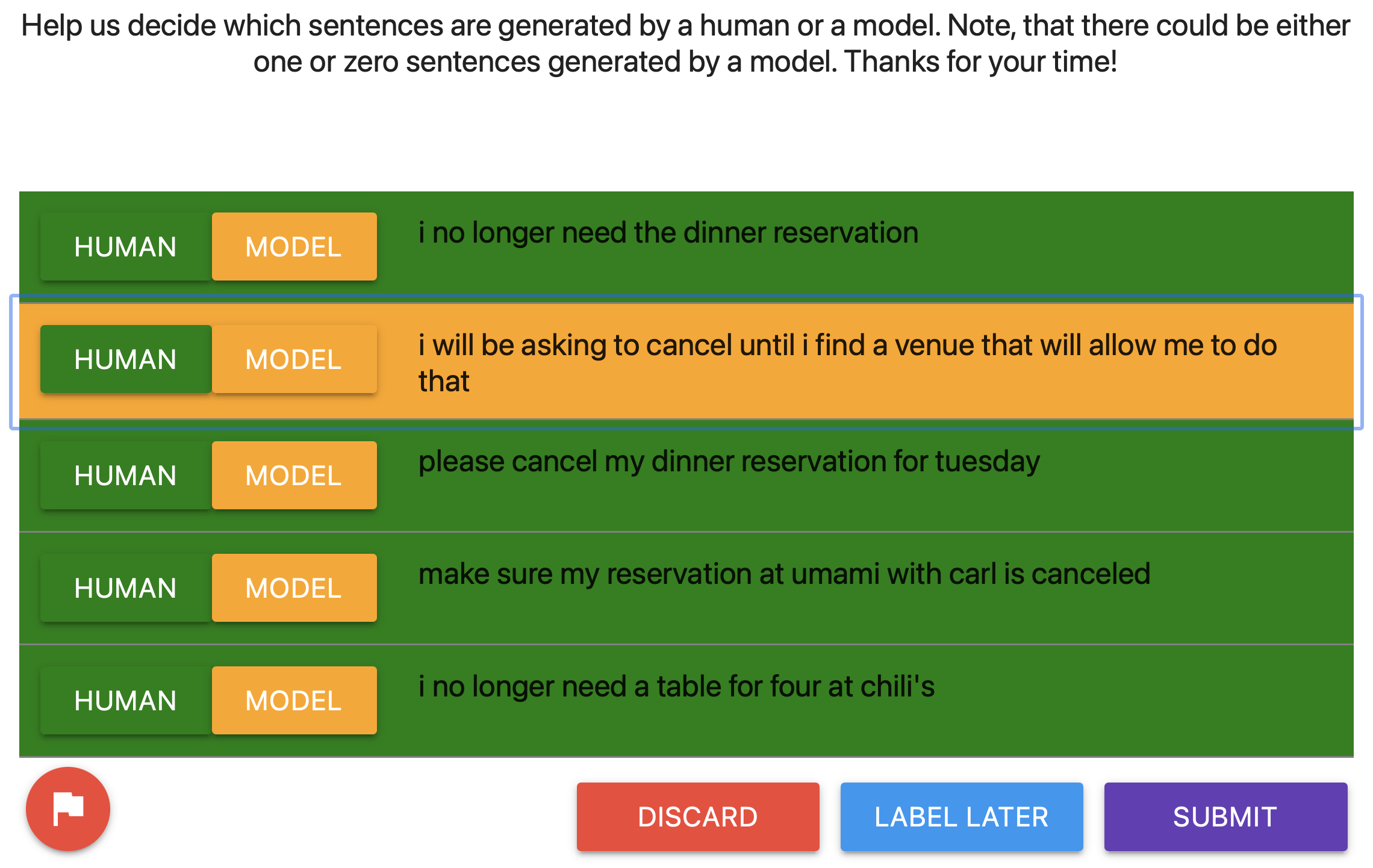}
    \caption{\textbf{Human evaluation tool.} Example of a question for the human evaluators. Human evaluators are asked to flag which example is GPT-3 generated if any among the 5 presented ones. }
    \label{fig:evaluator}
\end{figure}
We consider that a model produces sentences with high fidelity if a human is unable to distinguish them from a set of human-generated sentences belonging to the same intent. Therefore, for each intent in the CLINC150 dataset, we sample five random examples and we randomly choose whether to replace one of them by a GPT-3 generated sentence from the same intent. We generate sentences with each of the four GPT-3 models considered in the main text with two different temperatures (0.8 and 1.0). The sentence to replace is randomly selected. Finally, the five sentences are displayed to a human who has to choose which of the sentences is generated by GPT-3, if any. 

The task is presented to human evaluators in the form of a web application (see Figure~\ref{fig:evaluator}). We placed a button next to each sentence in order to force human evaluators to individually consider each of the examples. Once annotated, the evaluator can either \textit{submit}, \textit{discard}, or leave the task to \textit{label later}. We used a set of 15 voluntary evaluators from multiple backgrounds, nationalities, and genders. Each evaluator annotated an average of 35 examples, reaching a total of 500 evaluated tasks.

For each model and temperature, we report the error rate of humans evaluating whether a task contains a GPT-generated sample. We consider that evaluators succeeds at a given task when they correctly find the sentence that was generated by GPT or when they identify that none of them was generated. Thus, the error rate for a given model and temperature is calculated as \#failed / total\_evaluated.

Results are displayed in Figure~\ref{fig:human_evaluation}. We find that human evaluators tend to make more mistakes when the temperature used to sample sentences from GPT-3 is smaller. This result is expected since lowering the temperature results in sentences closer to those prompted to GPT-3, which are human-made. We also observe that models with higher capacity such as \texttt{Davinci} tend to generate more indistinguishable sentences than lower-capacity models such as \texttt{Ada}, even for higher temperatures. These results are in agreement with the "oracle" fidelity results introduced in Figure~\ref{fig:temperatures}.
\end{document}